# Improved Inception-Residual Convolutional Neural Network for Object Recognition

Md Zahangir Alom[1*], Mahmudul Hasan[2], Chris Yakopcic[1], Tarek M. Taha[1], and Vijayan K. Asari[1]
[1]Department of Electrical and Computer Engineering, University of Dayton, OH, USA
[2]Comcast Labs, Washington, DC, USA
Emails: [1*]alomm1@udayton.edu, [2]mahmud.ucr@gmail.com, [1]{cyakopcic1, ttaha1, vasari1}@udayton.edu

**Abstract**
Machine learning and computer vision have driven many of the greatest advances in the modeling of Deep Convolutional Neural Networks (DCNNs). Nowadays, most of the research has been focused on improving recognition accuracy with better DCNN models and learning approaches. The recurrent convolutional approach is not applied very much, other than in a few DCNN architectures. On the other hand, Inception-v4 and Residual networks have promptly become popular among computer the vision community. In this paper, we introduce a new DCNN model called the Inception Recurrent Residual Convolutional Neural Network (IRRCNN), which utilizes the power of the Recurrent Convolutional Neural Network (RCNN), the Inception network, and the Residual network. This approach improves the recognition accuracy of the Inception-residual network with same number of network parameters. In addition, this proposed architecture generalizes the Inception network, the RCNN, and the Residual network with significantly improved training accuracy. We have empirically evaluated the performance of the IRRCNN model on different benchmarks including CIFAR-10, CIFAR-100, TinyImageNet-200, and CU3D-100. The experimental results show higher recognition accuracy against most of the popular DCNN models including the RCNN. We have also investigated the performance of the IRRCNN approach against the Equivalent Inception Network (EIN) and the Equivalent Inception Residual Network (EIRN) counterpart on the CIFAR-100 dataset. We report around 4.53%, 4.49% and 3.56% improvement in classification accuracy compared with the RCNN, EIN, and EIRN on the CIFAR-100 dataset respectively. Furthermore, the experiment has been conducted on the TinyImageNet-200 and CU3D-100 datasets where the IRRCNN provides better testing accuracy compared to the Inception Recurrent CNN (IRCNN), the EIN, and the EIRN.

## I. Introduction

Recently, deep learning using Convolutional Neural Networks (CNNs) has shown great success in the field of machine learning and computer vision. The CNNs provide state-of-the-art accuracy in various image recognition tasks including object recognition [26], segmentation [1], human activity analysis [2], image super resolution [3], object detection [5], tracking [6], image captioning [7], and scene understanding [4, 5]. Additionally, this approach has been applied passively in video processing tasks including video classification [8], video representation, and classification of human activity [10]. Deep learning is applied in sentiment analysis which is used for online movie recommendation systems, in addition to other applications [8]. Deep learning approaches are used in the field of machine translation and natural language understanding, and they achieve state-of-the accuracy in this application domain [11, 12]. Furthermore, this technique has been used extensively in the field of speech recognition [13]. Moreover, the deep learning technique is not limited to signal, natural language, image, and video processing tasks; it has been successfully applied in the field of game development [14, 15].

Machine intelligence provides improved performance in many different fields including calculation, chess, memory, and pattern matching, where as human intelligence still shows better performance in the fields of object recognition and scene understanding tasks. In the recent years, deep learning techniques (DCNNs in particular) have been providing outstanding performance for most of the tasks in computer vision. The DCNN is a hierarchical feature learning approach with multi-level and multi-scale abstraction of features, which aids in the learning of global contextual information from the input samples. However, there is still a gap that must be closed before human level intelligence can be achieved when performing visual recognition tasks. To reach human level performance during recognition tasks, a lot of research is dedicated to understanding the actual process of recognition, as well as understanding the tasks of the visual cortex in the human brain. Studies show that the human brain processes visual information using operations that are similar to convolution or filtering, activation, pooling, and normalization with recurrent connectivity in the visual cortex [16]. The recurrent connectivity of synapses in the human brain plays a big role for context modeling in visual recognition tasks [16, 17].

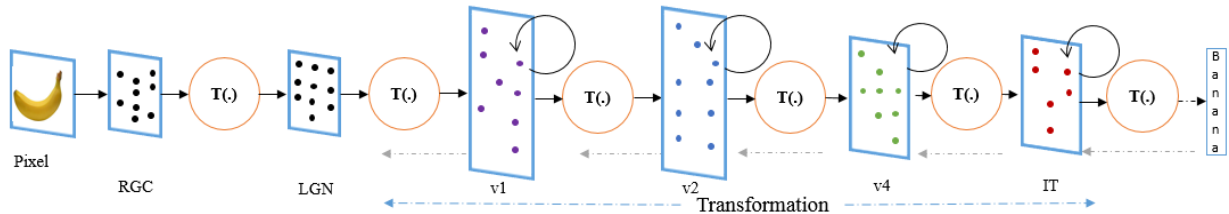

**Figure 1.** Visual information processing pipeline of the human brain, where v1 though v4 represent the visual cortex areas. The visual context areas of v1 though v4 process information using recurrent techniques.

If we observe the structure of recently developed DCNN models, most functionalities are included to design better architectures which are successfully applied to segmentation, detection, and recognition tasks. However, the concept of Recurrent Convolution Layers (RCLs) is included in very few DCNN models, the most prominent being the Recurrent Convolutional Neural Network (RCNN) [18], a CNN with LSTM for object classification [19], and the Inception RCNN [20]. On the other hand, Inception [22], and Residual [21, 23] architectures are commonly used for solving computer vision tasks. The common practice in the most recently developed Inception and Residual networks is to implement larger and deeper networks to archive better performance. As the model becomes larger and deeper, the parameters of the network are increased dramatically. As a result, the model becomes more complex to train and thus, more computationally expensive. Therefore, it is very important to design an architecture which provides better performance using reasonably fewer numbers of network parameters.

While others are trying to implement bigger and deeper DCNN architectures like GoogLeNet [24], or a Residual Network with 1001 layers [21] to achieve high recognition accuracy on different benchmark datasets. We are presenting an improved version of the DCNN model inspired by the recently developed promising DCNN architectures like Inception-v4 [22], Residual [23], and the RCNN [18]. The proposed model not only ensures better recognition accuracy with same number of network parameters against other DCNN architectures, but also helps to improve the overall training accuracy. The contributions of this work are as follows:

- A new deep learning model named the Inception Recurrent Residual Convolutional Neural Network (IR RCNN) is proposed.
- Empirical evaluation of the performance of the proposed model against different DCNN models on different benchmark datasets such as CIFAR-10, CIFAR-100, TinyImageNet-200, and CU3D-100.
- Empirical investigation of the impact of the RCLs of the IRRCNN against that of the equivalent Inception and Inception-Residual models on the CIFAR-100 and TinyImageNet-200 datasets.
- Large scale implementation and comparison against Inception-v3 on the CU3D-100 object recognition dataset.

The rest of the paper has been organized as follows: Section II describes related work, and Section III presents the theoretical details of the IRRCNN model. Results and discussion are provided in Section IV, and conclusions and future work are discussed in Section V.

## II. Related work

Most of the breakthroughs in the field of computer vision (as well as the ImageNet challenges) have driven the development of the different DCNN architectures in recent years. The deep learning revolution began in 1998 with [25]. From then on, several different architectures have been proposed that have shown great success using many different benchmark datasets including MNIST, SVHN, CIFAR-10, CIFAR-100, ImageNet, and many more. Of the DCNN architectures, AlexNet [26], VGG [27], NiN [28], the All Convolutional Network [29], GoogLeNet [24], Inception-v4 [22], and the Residual Network [23] can be considered the most popular deep learning architectures due to their outstanding performance on different benchmarks for object classification tasks. In most cases, researchers experiment with different models such as NIN, the All Convolutional Network, VGG, GoogLeNet, Inception, and Residual networks, and then select the best model for their application based on the performance. Nevertheless, new models, hybrid models, and optimized versions of existing models have been proposed to achieve better accuracy with less network parameters in the last few years.

The concept of Inception was introduced with GoogLeNet [24], and it won the most difficult ImageNet challenge for visual object recognition called the ImageNet Large Scale Visual Recognition Challenge (ILSVRC) in 2014 with remarkably few parameters. The main contribution of this network is to reduce the network parameters drastically when compared to the traditional CNN used in AlexNet. This model introduced a new technique called an inception layer. This approach is not only computationally convenient when compared to the traditional approach, but it also provided the best recognition accuracy in ILSVRC 2014. In terms of network parameters and memory, GoogLeNet needs only 4M whereas AlexNet needs around 60M [24]. An improved version of the Inception network was proposed by Szegedy al et. in 2015, where they scaled up the Inception model utilizing more computation with factorized convolution and aggressive regularization [22,30]. This model shows a significant improvement in recognition accuracy on ILSVRC 2012.

In 2015, Kiming He al et. proposed a new DCNN architecture called the Residual Network [23] and won the most difficult ILSVRC in 2015. This deep learning technique achieves state-of-the-art recognition accuracy on different benchmarks including ImageNet and CIFAR, as well as on object detection and segmentation tasks on PASCAL VOC and MSCOCO. This architecture is applied to different application domains including machine translation [31], speech synthesis [32], speech recognition [33] and audio classification [34]. Residual networks provide the possibility of building deep network architectures with thousands of layers resulting in significantly improved recognition accuracy [21]. However, improving just a fraction of a percentage in recognition accuracy requires almost doubling the number of layers in the networks. As a result, the number of model parameters and complexity increases. Therefore, training with very deep networks becomes very difficult due to diminishing feature reuse, which makes the networks very slow to train. Research has been conducted focusing on designing alternative models that produce the same level of recognition accuracy like SqueezeNet, which requires significantly less model parameters [37]. To overcome the problem of training complexity in residual networks, wide residual networks (WRN) have been proposed [35], where the width (number of feature maps) of the networks is increased instead of the depth (number of layers). In 2016, the aggregated residual network was also proposed which is a slight variant of the basic residual network structure [36].

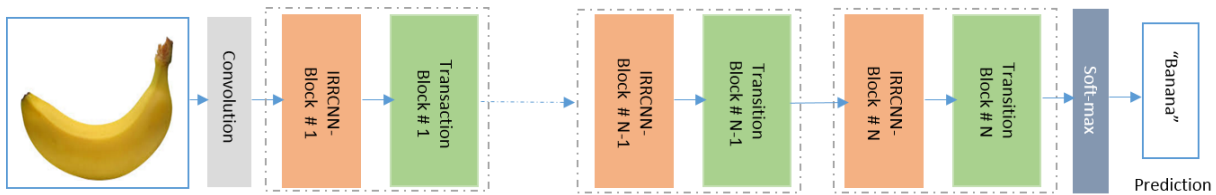

**Figure 2.** The overall layer flow diagram of proposed IRRCNN) consisting of the IRRCNN-Block, the IRRCNN-Transition block, and the Softmax layer at the end.

Most existing research has been concentrated on improving recognition accuracy with different DCNN models. Out of the many modes, very few studies are using RCLs in their models. However, the recurrent approach is very important for context modeling in sequential images and videos. The RCNN structure was proposed for object recognition tasks by Ming et al in 2015 [18]. This deep learning model contains several blocks of RCLs followed by a max-pooling layer. The global max-pooling layer is placed before the classification layer with Softmax at the end. This model provided state-of-the-art accuracy for object classification at that time [18]. In 2014, the Long-term Recurrent Convolutional Network (LRCN) was proposed for visual recognition and description by Donahue et al. [19]. This architecture contains of two popular techniques, the CNN and LSTM. The CNN technique is used for feature extraction, and LSTM is applied to observe how features vary with respect to time. This model shows outstanding performance for visual description [19]. Moreover, some research is being conducted that emphasizes on bridging gap between machine and human intelligence, where the proposed networks utilize recurrent concepts using residual network models [38].

Inception and Residual architectures are very prevalent in the computer vision community. The success of both architectures in the last few years has produced a new path of research that focuses on the discovery of even better models with better performance. Incorporating the new functionalities of RCLs into these state-of-the-art models improves overall recognition accuracy while utilizing the same number of the network parameters. This will have significant impact on the both computer vision and machine learning communities. In this paper, we have proposed an improved DCNN architecture based on Inception [22], Residual networks [23] and the RCNN architecture [18]. Therefore, we call this model the **Inception Recurrent Residual Convolutional Neural Network (IRRCNN)**.

### III. IRRCNN architecture

The main objective of this model is to improve recognition performance using the same number or fewer computational parameters when compared to alternative equivalent deep learning approaches. In this model, the inception-residual units utilized are based on Inception-v4 [22]. The Inception-v4 network created by Szeged al et. in 2015 is a deep learning model that concatenates the outputs of the convolution operations with different sized convolution kernels in the inception block [22]. Inception-v4 is a simplified structure of Inception-v3 containing more inception modules using lower rank filters. Furthermore, Incpetion-v4 includes a residual concept in the inception network called the Inception-v4 Residual Network, which improves overall accuracy of recognition tasks. In the Inception-Residual network, the outputs of the inception units are added to the inputs of the respective units. The overall structure of the proposed IRRCNN model is shown in Figure 2. From the figure, it can be clearly seen, that the overall model consists of several convolution layers, IRRCNN blocks, transition blocks, and a Softmax at the output layer.

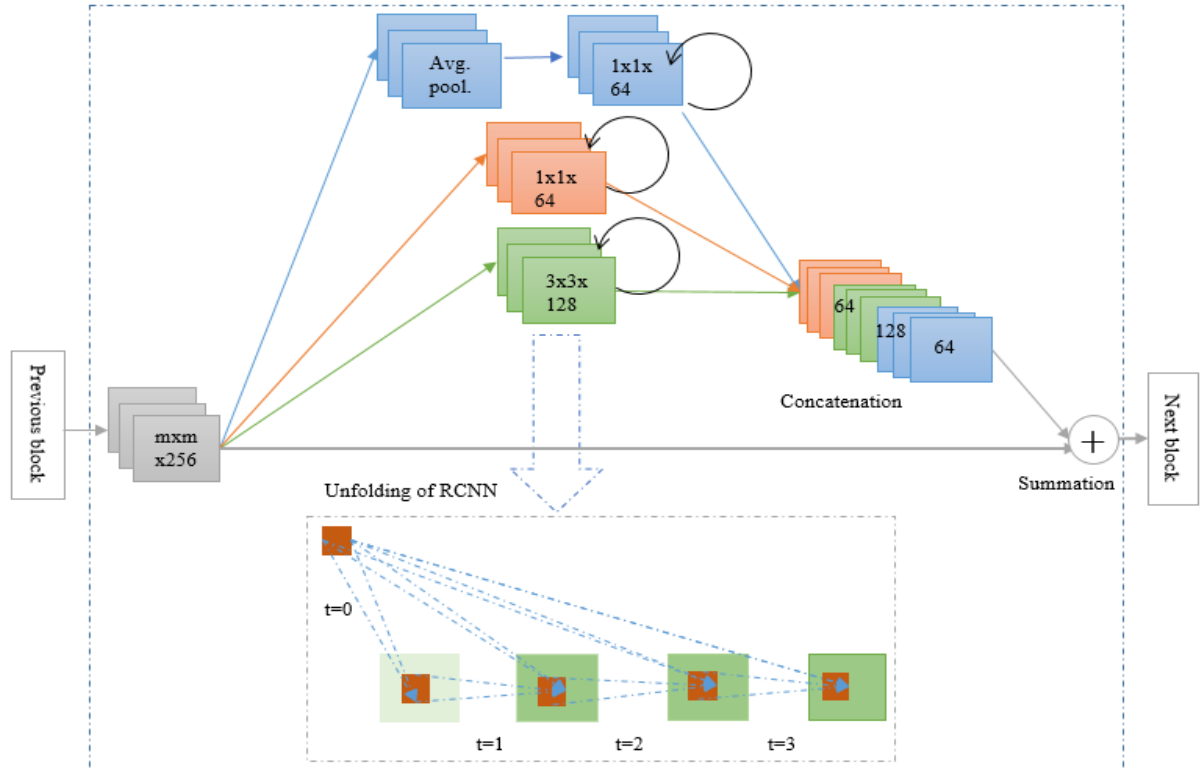

**Figure 3**. The Inception Recurrent Residual Convolutional Neural Network (IRRCNN) block consisting of the inception unit at the top which contains recurrent convolutional layers that are merged by concatenation, and the residual units (summation of the input features with the outputs of the inception unit can be seen at the end of the block).

The most significant part of this proposed architecture is the IRRCNN block that includes RCLs, inception units, and residual units (shown in detail in Figure 3). The inputs are fed into the input layer, then passed through inception units where RCLs are applied, and finally the outputs of the inception units are added to the inputs of the IRRCNN-block. The recurrent convolution operations perform with respect to the different sized kernels in the inception unit. Due to the recurrent structure within the convolution layer, the outputs at the present time step are added with the outputs of previous time step. The outputs at the present time step are then used as inputs for the next time step. The same operations are performed with respect to the time steps that are considered. For example, here *k=2* means that 3 RCLs are included in IRRCNN-block. In the IRRCNN-block, the input and output dimensions do not change, this is simply an accumulation of feature maps with respect to the time steps. As a result, the healthier features ensure that better recognition accuracy is achieved with the same number of network parameters.

The operations of the RCL are performed with respect to the discrete time steps that are expressed according to the RCNN [19]. Let's consider the $x_l$ input sample in the $l^{th}$ layer of the IRRCNN-block and a pixel located at $(i,j)$ in an input sample on the $k^{th}$ feature map in the RCL. Additionally, let's assume the output of the network $O_{ijk}^l(t)$ is at the time step $t$. The output can be expressed as follows:

$$O_{ijk}^l(t) = \left(w_k^f\right)^T * x_l^{f(i,j)}(t) + (w_k^r)^T * x_l^{r(i,j)}(t-1) + b_k \tag{1}$$

Here $x_l^{f(i,j)}(t)$ and $x_l^{r(i,j)}(t-1)$ are the inputs for the standard convolution layers and for the $l^{th}$ RCL respectively. The $w_k^f$ and $w_k^r$ values are the weights for the standard convolutional layer and the RCL of the $k^{th}$ feature map respectively, and $b_k$ is the bias.

$$y = f(O_{ijk}^l(t)) = \max(0, O_{ijk}^l(t)) \tag{2}$$

Here $f$ is the standard Rectified Linear Unit (ReLU) activation function. We have also explored the performance of this model with the Exponential Linear Unit (ELU) activation function in the following experiments. The outputs $y$ of the inception units for the different size kernels and average pooling layer are defined as $y_{1x1}(x), y_{3x3}(x)$, and $y_{1x1}^p(x)$ respectively. The final outputs of Inception Recurrent Convolutional Neural Networks (IRCNN) unit are defined as $\mathcal{F}(x_l, w_l)$ which can be expressed as

$$\mathcal{F}(x_l, w_l) = y_{1x1}(x) \odot y(x) \odot y_{1x1}^p(x) \tag{3}$$

Here $\odot$ represents the concatenation operation with respect to the channel or feature map axis. The outputs of the IRCNN-unit are then added with the inputs of the IRRCNN-block. The residual operation of the IRRCNN-block can be expressed by the following equation.

$$x_{l+1} = x_l + \mathcal{F}(x_l, w_l) \tag{4}$$

Where $x_{l+1}$ refers to the inputs for the immediate next transition block, $x_l$ represents the input samples of the IRRCNN-block, $w_l$ represents the kernel weights of the $l^{th}$ IRRCNN-block, and $\mathcal{F}(x_l, w_l)$ represents the outputs from of $l^{th}$ layer of the IRCNN-unit. However, the number of feature maps and the dimensions of the feature maps for the residual units are the same as in the IRRCNN-block shown in Figure 3. Batch normalization is applied to the outputs of the IRRCNN-block [53]. Eventually, the outputs of this IRRCNN-block are fed to the inputs of the immediate next transition block.

In the **transition block**, different operations are performed including convolution, pooling, and dropout, depending upon the placement of the transition block in the network. We did not include inception units in the transition block on the small-scale implementation for CIFAR-10 and CIFAR-100. However, we have applied inception units to the transition block during the experiment using the TinyImageNet-200 dataset and for the large-scale model which is the equivalent model of Inception-v3 [30]. The down-sampling operations are performed in the transition block where we perform max-pooling operations with a 3×3 patch and a 2×2 stride. The non-overlapping max-pooling operation has a negative impact on model regularization, therefore we used overlapped max-pooling for regularizing the network which is very important when training a deep network architecture [24]. Late use of a pooling layer helps to increase the non-linearity of the features in the network, as this results in higher dimensional feature maps being passed through the convolution layers in the network. We have applied two special pooling layers in the model with three IRRCNN-blocks and a transition-block for the experiments that use the CIFAR-10 or CIFAR-100 dataset.

We used only 1×1 and 3×3 convolution filters in this implementation, as inspired by the NiN [28] and Squeeze Net [37] models. This also helps to keep the number of network parameters at a minimum. The benefit of adding a 1×1 filter is that it helps to increase the non-linearity of the decision function without having any impact on the convolution layer. Since the size of the input and output features does not change in the IRRCNN blocks, it is just a linear projection on the same dimension and non-linearity is added to the RELU and ELU activation functions. We used a 0.5 dropout after each convolution layer in the transition block. Finally, we used a Softmax, or normalized exponential function layer at the end of the architecture. For input sample $x$, weight vector $W$, and $K$ distinct linear functions, the Softmax operation can be defined for the $i^{th}$ class as follows:

$$P(y=i|x) = \frac{e^{x^T w_i}}{\sum_{k=1}^{K} e^{x^T w_k}} \tag{5}$$

This proposed IRRCNN model has been investigated through a set of experiments on different benchmark datasets and compared across different models.

## IV. Experiments

The proposed IRRCNN model has been evaluated using four different benchmark datasets: CIFAR-10[39.], CIFAR-100[39], TinyImageNet-200 [40], and CU3D–100 [41]. The dataset statistics are provided in Table 1. We used different validation and testing samples for the TinyImageNet-200 dataset. The entire experiment was conducted on a Linux environment with Keras [54] and Theano [55] at the backend running on a single GPU machine with an NVIDIA GTX-980Ti.

**Table 1.** Statistics for the datasets studied in these experiments.

| Dataset | Training Samples | Validation/Testing Samples | Total Samples |
| --- | --- | --- | --- |
| CIFAR-10 | 50,000 | 10,000/10,000 (same) | 60,000 |
| CIFAR-00 | 50,000 | 10,000/10,000 (same) | 60,000 |
| TinyImageNet-200 | 100,000 | 10,000/10,000 (different) | 120,000 |
| CU3D-100 | 14,130 | 4,710/4,710 (same) | 18,840 |

### 4.1 Experiments on CIFAR-10 and 100 datasets

In this experiment, we used two convolution layers at the beginning of the architecture, three IRRCNN blocks followed by three transition blocks, and one global average pooling and Softmax layer at the end. First, we evaluated the IRRCNN model using the stochastic gradient descent (SGD) technique with the Keras 2.0 [55] default initialization technique. We used momentum equal to 0.9 [42] and decay equal to 9.99e-07 in this experiment. Second, we evaluated the same model with the Layer-sequential unit-variance (LSUV) initialization method [43] and the latest improved version of the optimization function called EVE [44]. The hyper parameters for the EVE optimization function are as follows: the value of learning rate ($\lambda$) is 1e-4, the decay ($\gamma$) is 1e-4, $\beta_1 = 0.9$, $\beta_2 = 0.999$, $\beta_3 = 0.999b$, $\kappa=0.1$, K=10, and $\epsilon = 1e-08$. The values $\beta_1, \beta_2 \in [0,1)$ are exponential decay rates for moment estimation in Adam. The $\beta_3 \in [0,1)$ is exponential decay rate for computing relative changes. The IRRCNN-block uses the $l2-norm$ for a weight regularization of 0.002. We used the ReLU activation function in the first experiment, and the ELU activation is used in the second experiment. In both experiments, we trained the networks for 350 epochs with a batch size of 128 for CIFAR-10 and 100.

**CIFAR -10:** The CIFAR-10 dataset is a benchmark dataset for object classification [39]. The dataset consists of 32×32 color images split into 50,000 samples for training, and the remaining 10,000 samples are used for testing (classification into one of 10 classes). The experiment was conducted with and without data augmentation. When using data augmentation, we applied only random horizontal flipping. Using this proposed approach, we have achieved around 8.41% testing error without data augmentation and 7.37% testing error with augmented data (only horizontal flipping) using SDG techniques.

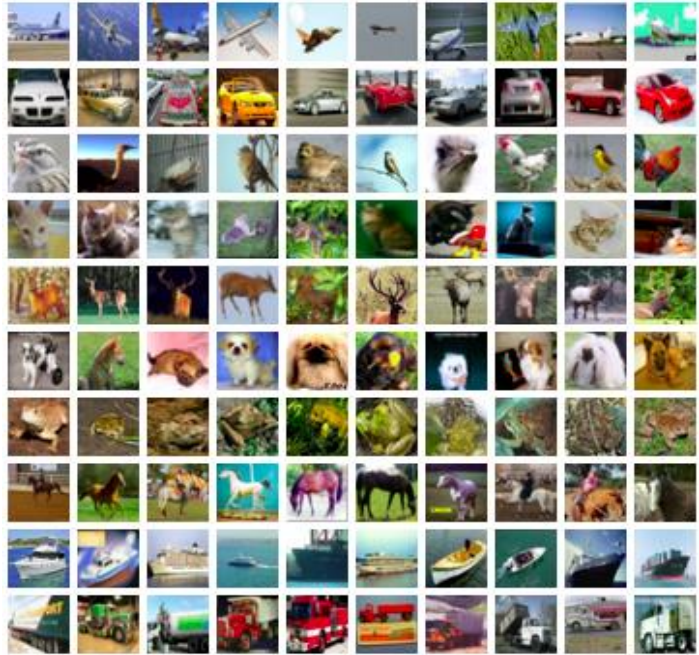

**Figure 4.** Example images from the CIFAR-10 dataset.

The proposed model shows better recognition against most of the DCNN models displayed in Table 2. Furthermore, improved performance is observed in the IRCNN that used LSUV [43] initialization and the EVE [44] optimization function. The results show a testing error of around 8.17% and 7.11% without and with data augmentation respectively. It is also observed that the IRRCNN shows better performance when compared to the equivalent IRCNN model [20].

**Table 2.** Testing error (%) of the IRRCNN on CIFAR-10 object classification dataset without and with data augmentation. For unbiased comparison, we have listed the accuracy stated in recent studies using a similar experimental setting.

| Methods | # Parameters | Error (%) without data augmentation | Error (%) with data augmentation |
|---|---|---|---|
| Maxout [ 46] | >6M | 11.6 | 9.38 |
| Network in Network (NiN) [28] | ~1M | 10.4 | 8.81 |
| Deeply Supervised Network DSN [ 47] | ~1M | 9.69 | 7.97 |
| Conv. Net + Probout [48] | | 9.39 | - |
| ALL-CNN [ 29] | | 9.08 | 7.25 |
| Drop-Connect [ 49] | | - | 9.32 |
| Highway Network [50] | | - | 7.72 |
| RCNN [18] | | 8.69 | 7.09 |
| dasNet[51] | | - | 9.22 |
| FitNet [ 45] | ~2.5M | - | 8.39 |
| Residual Net [23] | | | 7.51 |
| BO using DNN[ 52] | | | 6.37 |
| IRCNN +SDG+ReLU | 3.5 M | 8.41 | 7.37 |
| IRCNN + LSUV + EVE +ReLU | 3.5 M | 8.17 | 7.11 |
| IRRCNN+SGD+ReLU | 3.5 M | 8.14 | 7.23 |
| IRRCNN + LSUV + EVE+ReLU | 3.5 M | 8.11 | **7.06** |

**CIFAR-100:** Another similar benchmark for object classification was developed in 2009 [39]. The dataset contains 50,000 samples for training and 10,000 samples for validation and testing. Each sample is a 32×32×3 image, and the dataset has 100 classes. The proposed IRRCNN model was studied with and without data augmentation. During the experiment with augmented data, the SGD and LSUV [43] initialization approaches and the EVE optimization function were used [44]. In both cases, the proposed technique shows better recognition accuracy compared with different DCNN models including the IRCNN [20]. The validation accuracy of the IRRCNN model for both experiments on CIFAR-100 with data augmentation is shown in Figure 5. The proposed IRRCNN model shows better performance in the both experiments when compared to the IRCNN [20], EIN, and EIRN models. The experimental results when using CIFAR-100 are shown in Table 3. The IRRCNN model provides better testing accuracy compared to many recently developed methods. We have achieved 72.78% recognition accuracy with LSUV+EVE which is around a 4.49% improvement compared to one of the baseline RCNN methods with almost the same number of parameters (~3.5M) [18].

**Table 3.** Testing error (%) of the IRRCNN on the CIFAR-100 object classification dataset without and with data augmentation. For unbiased comparison, we have listed the accuracy provided by recent studies in a similar experimental setting.

| Methods | # of parameters | Error (%) without data augmentation | Error (%) with data augmentation |
|---|---|---|---|
| CNN+Tree based priors [47] | - | | 36.85 |
| Maxout [40] | >5M | | 38.57 |
| Prob maxout [45] | >5M | | 38.14 |
| NIN [21] | 1.98M | 35.68 | 35.68 |
| DSN [43] | 1.98M | | 34.57 |
| RCNN-160 [19] | 1.87M | | 31.75 |
| dasNet [42] | - | | 33.78 |
| ALL Conv [22] | - | | 33.71 |
| HighwayNet[44] | - | | 32.24 |
| FitNet [46] | - | | 35.04 |
| BO using DNN [52] | | | 27.40 |
| IRCNN+SGD+ReLU | ~3.5M | 34.13 | 31.22 |
| IRCNN+LSUV+EVE+ReLU | ~3.5M | 30.87 | 28.24 |
| IRRCNN+SGD+ELU | ~3.5M | 33.07 | 29.21 |
| IRRCNN + LSUV + EVE+ELU | ~3.5M | **29.67** | **27.10** |

### 4.2 Impact of recurrent convolution layers

**A question may arise here:** is there any advantage of the IRRCNN model against the EIRN and EIN architectures? The EIN and EIRN models are implemented with a similar architecture with same number of network parameters (~3.5 M). We used sequential convolution layers with the same time-step with the same size kernels instead of using RCLs for implementing the EIN and EIRN models. In addition, in the case of EIRN, we incorporated the residual concept with an Inception-block like Inception-v4 [22]. Furthermore, we have investigated the performance of the IRRCNN model against the RCNN with same number of parameters on the TinyImageNet-200 dataset.

**A possible second question may arise:** Is the IRRCNN model providing better performance due to the use of advance deep learning techniques? It is noted that LSUV initialization approach applied to the DCNN architecture called FitNet4 achieved 70.04% classification accuracy on augmented data with mirroring and random shifts for CIFAR-100 [45]. In contrast, we only applied random horizontal flipping for data augmentation and achieved around 1.76% better recognition accuracy against FitNet4 [45].

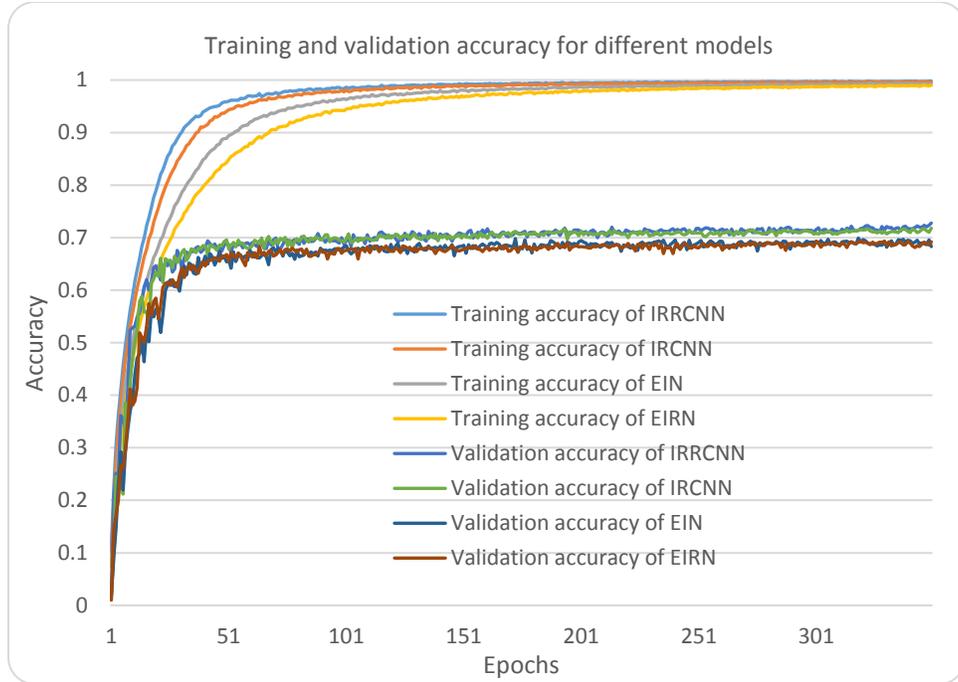

**Figure 5.** Training and validation accuracy for IRRCNN, IRCNN, BIN, and BIRN on CIFAR-100. The vertical and horizontal axis represents accuracy and epochs respectively. Our proposed model shows the best recognition accuracy in all cases.

The model accuracy for both training and validation are shown in Figure 5. From the figures, it is clearly observed that this proposed model shows lower loss and highest recognition accuracy compared to EIN and EIRN, which proves the necessity of the proposed models. The testing accuracy of IRRCNN, IRCNN, EIN, and EIRN are shown in Figure 6. It can be summarized that the proposed IRRCNN provides around 1.02%, 4.49%, and 3.56% improved testing accuracy compared to IRCNN, EIN, and EIRN respectively.

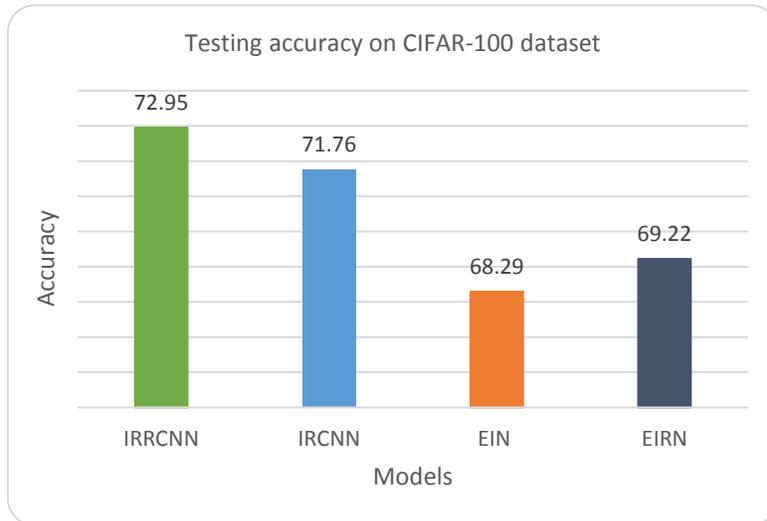

**Figure 6.** Testing accuracy of the proposed IRRCNN model against IRCNN, EIN, and EIRN on the augmented CIFAR-100 dataset.

### 4.3 Experiment on TinyImageNet-200

We also evaluated the proposed approach on the TinyImageNet-200 dataset [40]. This dataset contains 100,000 samples for training, 10,000 samples for validation, and 10,000 samples for testing. These images are sourced from

200 different classes of objects. The main difference between the main ImageNet dataset and Tiny ImageNet is the images are down sampled from 224x224 to 64x64. There are some negative impacts of down-sampling, like loss of detail. Therefore, down sampling the images leads to ambiguity, which makes this problem even harder and this effects overall model accuracy.

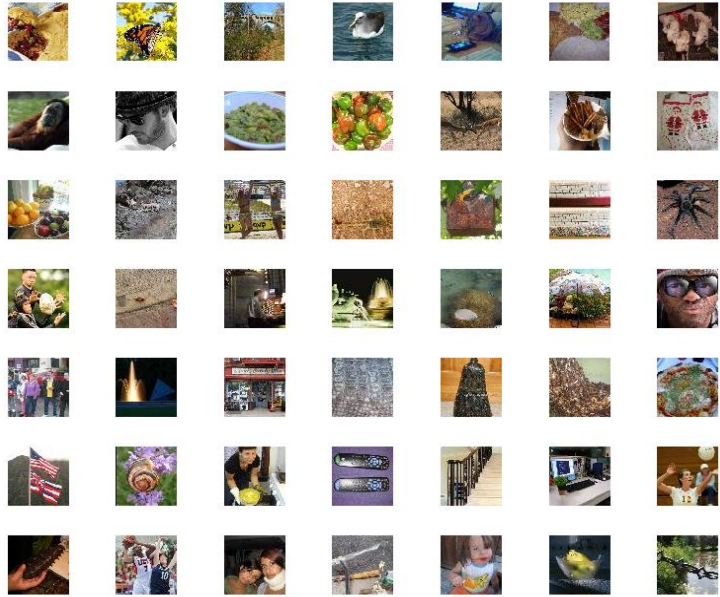

**Figure 7:** Sample images from the TinyImageNet-200 dataset.

For this experiment, we used the IRRCNN model with two general convolution layers with a 3×3 kernel at the beginning of the network followed by sub-sampling layer with 3×3 convolution using a stride of 2×2. After that, four IRRCNN blocks are used followed by four transition blocks. Finally, a global average pooling layer is used followed by a Softmax layer.

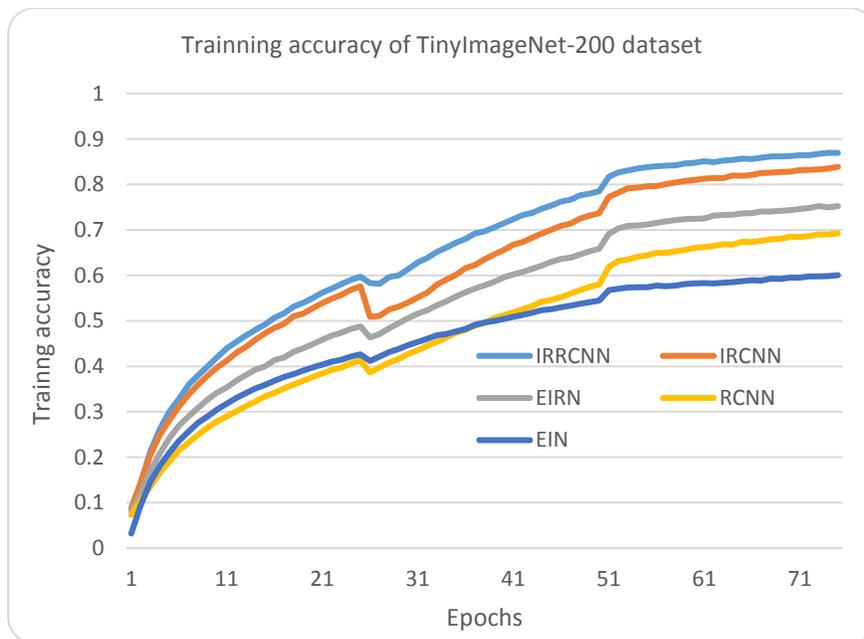

**Figure 8.** Training accuracy during training for TinyImageNet-200 dataset.

We have experimented with the IRRCNN, IRCNN, equivalent RCNN, EIN, and EIRN using the TinyImageNet-200 dataset. The training accuracy of this experiment is shown in Figure 8. The proposed IRRCNN model provides better recognition accuracy during training compared to equivalent models including IRCNN, EIN, and EIRN with almost the same number of network parameters (~15M). Generally, DCNN takes a lot of time and power when training a reasonably large model. The Inception-Residual networks with RCLs significantly reduce training time with faster convergence and better recognition accuracy. The validation accuracy for all of these models is shown in Figure 9.

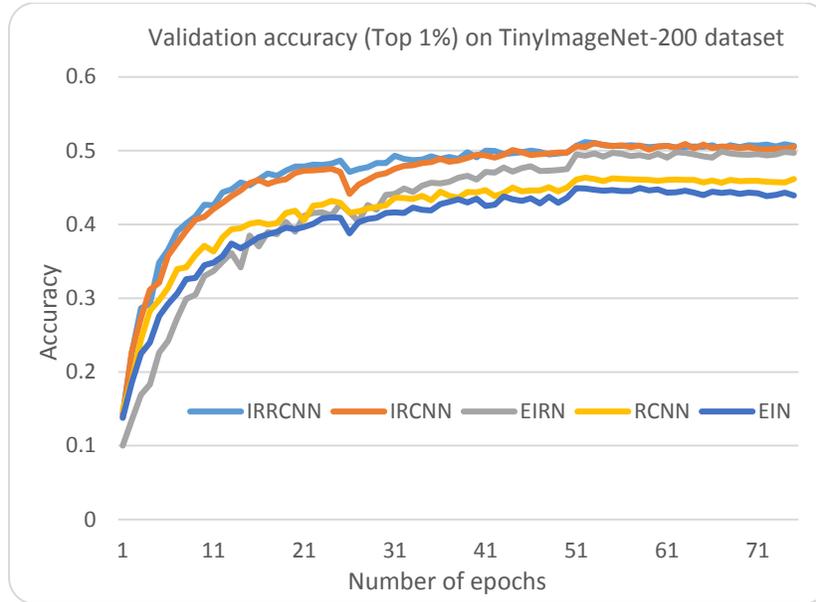

**Figure 9.** Validation accuracy on the Tiny-ImageNet dataset.

We have evaluated our proposed approach for both Top-1% and Top-5% testing accuracy as shown in Figure 10. From the bar graph, the impact of recurrent connectivity is clearly observed, and we have achieved 52.23% top-1% testing accuracy whereas the EIRN and EIN show 51.14% and 45.63% top-1% testing accuracy. The same behavior is observed for Top-5% accuracy as well. The IRRCNN provides better testing accuracy when compared against all other models in both cases which absolutely displays the robustness of the proposed deep learning architecture.

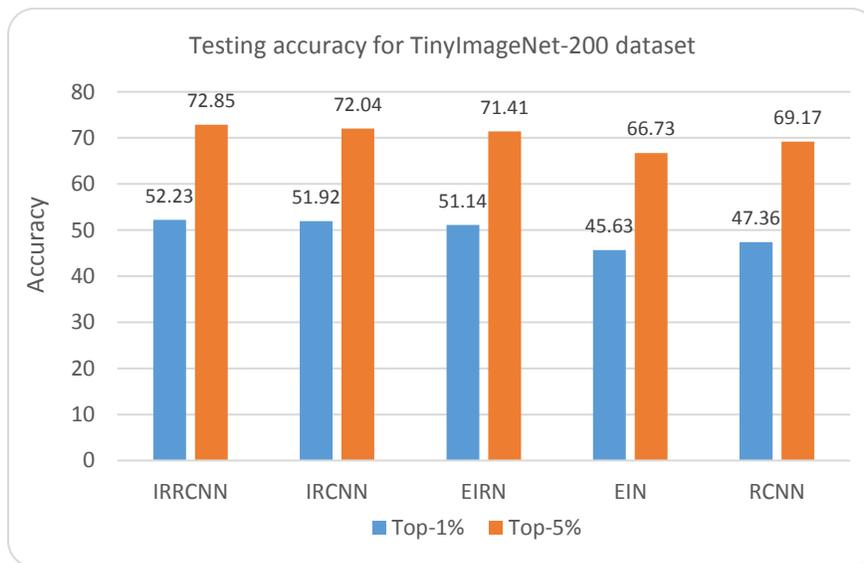

**Figure 10.** Top-1% and Top-5% testing accuracy on TinyImageNet-200 dataset.

### 4.4 Inception-v3 versus equivalent IRRCNN model

We evaluated the IRRCNN model with large scale implementation against the Ineption-v3 network for higher dimension input samples. The IRRCNN model is implemented a with similar structure to the Incpeiton-v3 for impartial comparison. We used the default implementation of Keras version 2.0 and we just incorporated the RCLs, where ($k = 2$), which means 3 RCLs are used in the Inception units and a residual layer is added at the end of the block. In this implementation, we initialized the Inception-v3 network with pretrained weights that were taken form github [56]. We trained the network with the SGD method with momentum. The concept of transfer learning is used where training was performed for 100 epochs in total. After successfully completing the initial training process for 50 epochs with learning rate of 0.001, the learned weights were used as initial weights for the next 50 epochs for fine-tuning of the network with a learning rate of 0.0001.

**CU3D-100 dataset:** Another very high quality visual object recognition dataset with well-controlled images (e.g., object invariance, features complexity) is CU3D-100, which is suitable for evaluation of new deep learning algorithms. This dataset contains 18,840 color images in total that have a dimension of 320×320×3 and 20 samples per exemplar. The following figure shows some example images from CU3D-100 dataset.

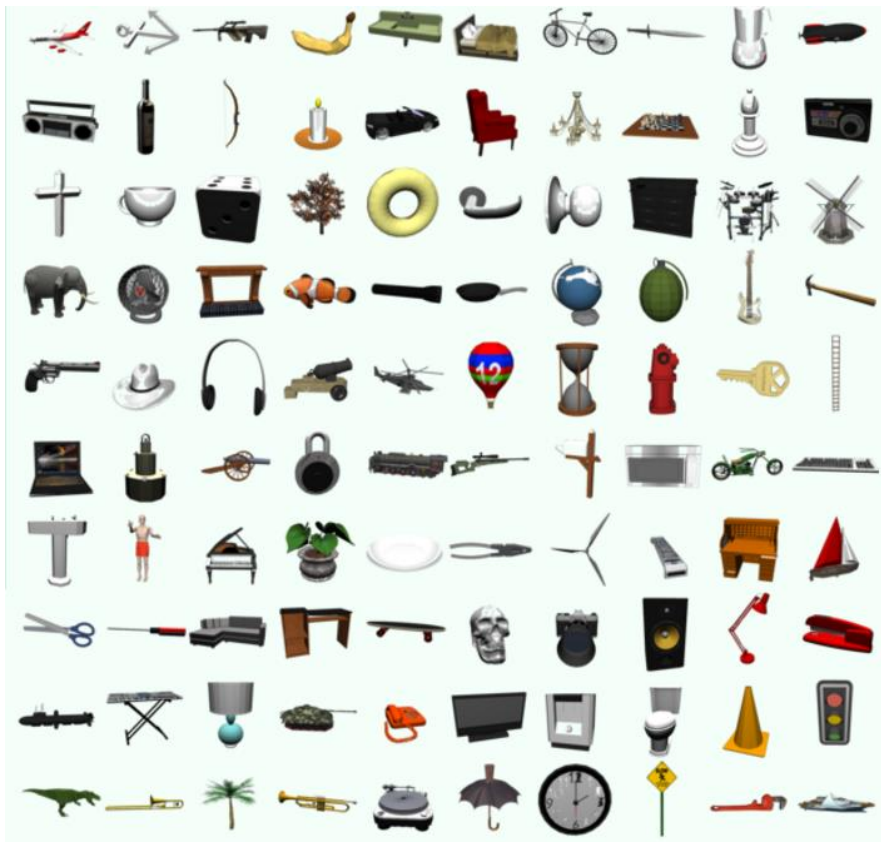

**Figure 11.** Example images of the CU3D-100 dataset.

The images in this dataset are three-dimensional views of real-world objects normalized for different positions, orientations, and scales. The rendered images have a $40^0$ depth rotation about the y-axis (plus a horizontal flip), a $20^0$ tilt rotation about x-axis, and an $80^0$ overhead lighting rotation. We used 75% percent of the images for training and the remaining 25% images for testing, which were selected randomly from the whole dataset. The example images in the fish category with different lighting condition and affine transformations are shown in Figure 12.

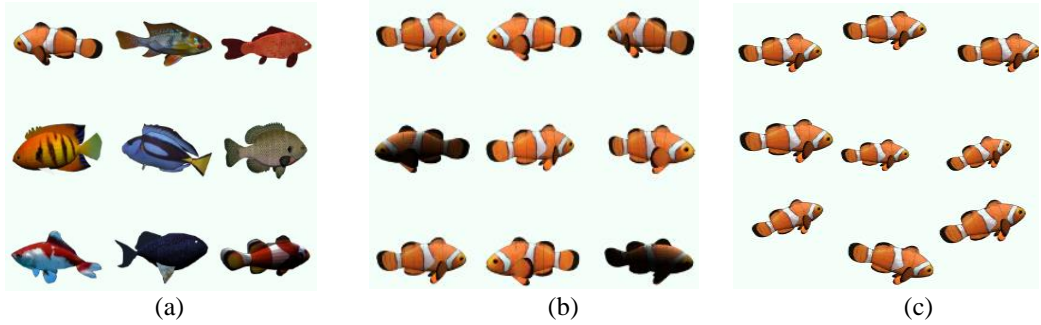

(a) (b) (c)

**Figure 12.** Sample images displaying (a) nine examples from the fish category, (b) nine depth, tilt, and lighting variations of the fish category, and (c) nine affine transformation images for a single view.

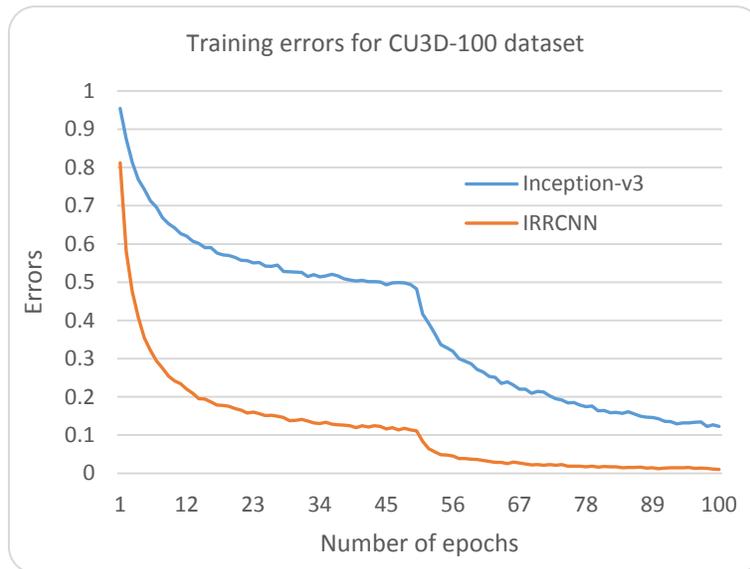

**Figure 13.** Training error with respect to the number of epochs on the CU3D-100 dataset

**Experimental results on CU3D**: The training errors and validation accuracy on the CU3D-100 dataset are shown in Figures 13 and 14 respectively. Figure 13 shows that the IRRCNN model exhibits lower error during training when compared to the Incpeiton-v3 model. Figure 14 shows the validation accuracy for 100 epochs; the experimental results show that our proposed approach gives better validation accuracy during training against the Inception-v3 model.

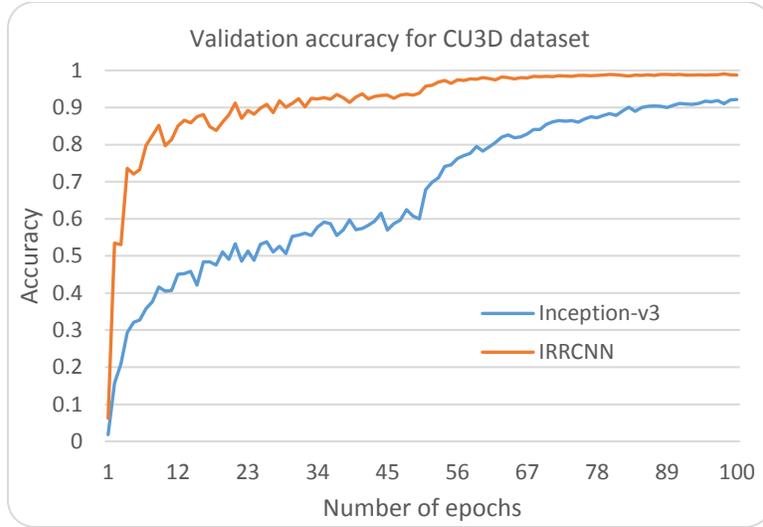

**Figure 14.** Validation accuracy with respect to epoch on the CU3D dataset

The proposed model gives about 98.84% testing accuracy on the CU3D dataset which is around 6.68% higher compared to the similar (both in structure and number of computational parameters) Inception-v3 model (see Figure 15). The total number of network parameters is about 23.5M for both models.

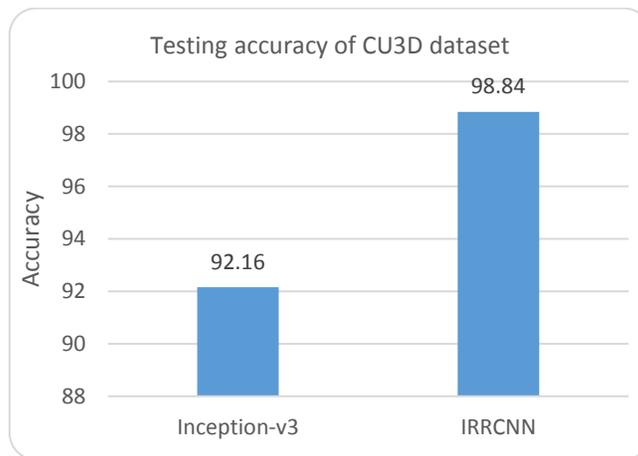

**Figure 15.** Testing accuracy of the IRRCNN model compared to that of the Inception-v3 network for the CU3D-100 dataset.

In addition, a recently published paper with sparse neural networks with recurrent layers reported about 94.6% testing accuracy on the CU3D dataset [41]. We have achieved around 4.24% better testing accuracy with the IRRCNN model. This experiment also proves the robustness of the IRRCNN model when dealing with scale invariance, position and rotation invariance, and different lighting condition input samples. The computational time of the proposed models as well as other equivalent models are shown in Table 4.

**Table 4.** Computational time for different models using different datasets.

| Models | Dataset | time/epoch (in sec.) |
|---|---|---|
| IRRCNN/IRCNN /EIN/EIRN | CIFAR-10 | ~422 |
| IRRCNN/IRCNN /EIN/EIRN | CIFAR-100 | ~425 |
| IRRCNN/IRCNN /EIN/EIRN | TinyImageNet-200 | ~765 |
| IRRCNN/ Inception-v3 | CU3D-100 | ~1858 |

## 4.5 Introspection

From our investigation, we have observed that the proposed IRRCNN model converges faster when compared to the RCNN, EIR, EIRN, and IRCNN models which are clearly evaluated using a set of experiments. The proposed techniques provide promising recognition accuracy during the testing phase with the same number of network parameters compared with other models. In this implementation, we have augmented input samples by applying only random horizontal flipping. From our observation, the proposed model will provide even better recognition accuracy with more augmentations including transition, central crop, and ZCA.

## V. Conclusion

In this paper, we have proposed the Inception Recurrent Residual Convolutional Neural Network (in short IRRCNN) for object recognition where we have utilized the power of recurrent convolution neural layers for context modulation based on the Inception and Residual Network architectures. The experimental results show promising recognition accuracy compared with different Deep Convolutional Neural Network (DCNN) models on different benchmarks including CIFAR-10, CIFAR-100, TinyImageNet-200, and CU3D-100. However, the proposed model has been evaluated with different advanced training approaches including SGD, initialization with Layer-sequential unit-variance (LSUV), and the recently proposed optimization methods of EVE. The IRRCNN model with LSUV and EVE achieved a promising object recognition accuracy of 72.78% on the CIFAR-100 dataset which is about a 4.53% improvement when compared to the Recurrent Convolutional Neural Network (RCNN) [19]. In addition, our model provides about 4.49% and 3.56% improvement in recognition accuracy when compared with Equivalent Inception Networks (EIN) and Equivalent Inception-Residual Networks (EIRN) on the CIFAR-100 dataset. Furthermore, we have achieved better recognition accuracy with IRRCNN when compared to EIRN, EIN, RCNN, and IRCNN on the TinyImageNet-200 dataset. Moreover, the large-scale implementation of the Inception-v3 network with recurrent convolutional layers (RCL) provides around 6.5% better recognition accuracy against the Incpetion-v3 model on the CU3D-100 dataset. Based on all experimental evaluations, it is clearly observed that the proposed architecture accelerates the training process, which is a big issue right now for training large scale deep learning approach. In the future, we would like to improve this model and explore segmentation and detection tasks.